\RequirePackage[svgnames]{xcolor}

\documentclass[11pt,letterpaper]{mystyle}

\usepackage{amsmath,amsfonts,bm,amsthm}

\def\eqref#1{equation~\ref{#1}}

\def\1{\bm{1}}

\DeclareMathAlphabet{\mathsfit}{\encodingdefault}{\sfdefault}{m}{sl}
\SetMathAlphabet{\mathsfit}{bold}{\encodingdefault}{\sfdefault}{bx}{n}

\newcommand{\R}{\mathbb{R}}

\DeclareMathOperator*{\argmin}{arg\,min}

\usepackage[all]{hypcap}
\usepackage[svgnames]{xcolor}
\usepackage[comma,authoryear,compress]{natbib}
\usepackage{hyperref}[citecolor=lightblue]

\hypersetup{
    colorlinks = true,
    citecolor = {teal},
}

\usepackage{algorithm}
\usepackage{algorithmicx}
\usepackage{algpseudocode}
\usepackage{microtype}
\usepackage{graphicx}
\expandafter\def\csname ver@subfig.sty\endcsname{}
\usepackage{booktabs} %
\usepackage{float}
\usepackage{bigstrut}
\usepackage{amsmath}
\usepackage{amssymb}
\usepackage{mathtools}
\usepackage{amsthm}
\usepackage{mathrsfs}
\usepackage{nicefrac}
\usepackage{dsfont}
\usepackage{enumitem}
\usepackage{subcaption}
\usepackage{graphicx,subfig}
\usepackage{cleveref}
\usepackage{bxcoloremoji}
\usepackage{datetime2}
\usepackage{float}
\usepackage{physics}
\theoremstyle{plain}
\newtheorem{theorem}{Theorem}[section]
\newtheorem{proposition}[theorem]{Proposition}

\theoremstyle{definition}

\theoremstyle{remark}
\newtheorem{remark}[theorem]{Remark}
\usepackage[utf8]{inputenc} %
\usepackage[T1]{fontenc}    %
\usepackage{hyperref}       %
\usepackage{url}            %
\usepackage{booktabs}       %
\usepackage{multirow}
\usepackage{amsfonts}       %
\usepackage{nicefrac}       %
\usepackage{microtype}      %
\usepackage{graphicx}
\usepackage{subcaption} 
\usepackage{wrapfig}
\usepackage{lipsum}
\usepackage{enumitem}
\usepackage{stackengine}
\usepackage[font=small,labelfont=bf]{caption}
\usepackage{color}
\usepackage{adjustbox}
\usepackage{multirow, multicol}

\usepackage{rotating}
\usepackage{makecell}

\usepackage{tcolorbox}
\tcbuselibrary{listings,skins,breakable}
\usepackage{listings}
\usepackage{subcaption}

\newcommand{\github}{\raisebox{-1.5pt}{\includegraphics[height=1.05em]{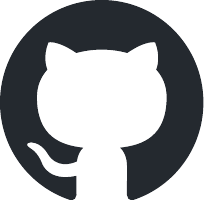}}}

\newtcolorbox{AIbox}[2][]{aibox,title=#2,#1}
\definecolor{lightgreen}{rgb}{0.22,0.70,0.30}%
\definecolor{Gray}{gray}{0.95}
\definecolor{Cornsilk}{rgb}{1.0, 0.97, 0.86}

\definecolor{lightblue}{HTML}{0064E0}
\definecolor{fg}{HTML}{1C2B33}
\definecolor{bg}{HTML}{F1F4F7}

\newcommand{\name}{\textsc{PEANuT}}

\usepackage{amsmath}

\usepackage[all]{hypcap}
\title{\LARGE PEANuT: Parameter-Efficient Adaptation with Weight-aware Neural Tweakers}

\runningtitle{PEANuT: Parameter-Efficient Adaptation with Weight-aware Neural Tweakers}

\author{
Yibo Zhong\textsuperscript{1,*} \quad
Haoxiang Jiang\textsuperscript{2,*} \quad
Lincan Li\textsuperscript{3} \quad
Ryumei Nakada\textsuperscript{4} \\
\bf Tianci Liu\textsuperscript{5} \quad
Linjun Zhang\textsuperscript{4} \quad
Huaxiu Yao\textsuperscript{6} \quad
Haoyu Wang\textsuperscript{2} \\
\textsuperscript{1}Independent Researcher \quad
\textsuperscript{2}University at Albany \quad
\textsuperscript{3}Florida State University \quad
\textsuperscript{4}Rutgers University \quad
\textsuperscript{5}Purdue University \quad
\textsuperscript{6}University of North Carolina at Chapel Hill
}

\begin{document}

\begin{abstract}
Fine-tuning large pre-trained foundation models often yields excellent downstream performance but is prohibitively expensive when updating all parameters. Parameter-efficient fine-tuning (PEFT) methods such as LoRA alleviate this by introducing lightweight update modules, yet they commonly rely on weight-agnostic linear approximations, limiting their expressiveness. In this work, we propose {\name}, a novel PEFT framework that introduces weight-aware neural tweakers, compact neural modules that generate task-adaptive updates conditioned on frozen pre-trained weights. {\name} provides a flexible yet efficient way to capture complex update patterns without full model tuning. We theoretically show that {\name} achieves equivalent or greater expressivity than existing linear PEFT methods with comparable or fewer parameters. Extensive experiments across four benchmarks with over twenty datasets demonstrate that {\name} consistently outperforms strong baselines in both NLP and vision tasks, while maintaining low computational overhead.
\vspace{2mm}


\vspace{5mm}

\coloremojicode{1F4C5} \textbf{Date}: \today


\github{} \textbf{Code Repository}: \href{https://github.com/yibozhong/peanut}{https://github.com/yibozhong/peanut} \



\coloremojicode{1F4E7} \textbf{Contact}: \href{mailto:yibozhong657@gmail.com}{yibozhong657@gmail.com}; \href{mailto:hjiang2@albany.edu}{hjiang2@albany.edu}; \href{mailto:hwang28@albany.edu}{hwang28@albany.edu}

\end{abstract}

\maketitle
\def\thefootnote{$^{*}$}\footnotetext{These authors contributed equally to this work, order was determined randomly (by rolling a die).}\def\thefootnote{\arabic{footnote}}
\vspace{3mm}
\vspace{-4mm}
\section{Introduction}
\vspace{-1mm}
Pre-trained models, trained on large and diverse general-domain corpora, have demonstrated strong generalization capabilities across a variety of tasks, including natural language understanding~\citep{devlin2018bert,liu2019roberta,howard2018universal,wu2019enriching,sun2023text,wang2021knowledge,wang2022fedkc}, generation~\citep{llama2-7b,llama3,xu2025collab,liu2025roserag,wang2024blendfilter,yao2022react,lewis2020retrieval}, and vision tasks such as image classification~\citep{dosovitskiy2020image,bhojanapalli2021understanding,chen2021crossvit}. A common strategy for adapting these models to specific downstream tasks is full fine-tuning. However, due to the massive number of parameters involved, full fine-tuning often leads to significant computational and memory costs~\citep{qin2024empirical}.

To mitigate these challenges, various parameter-efficient fine-tuning (PEFT) methods~\citep{ding2023parameter,han2024parameter} have been developed, enabling pre-trained models to be fine-tuned in resource-constrained environments~\citep{lin2024awq}. These methods retain most of the pre-trained weights in a frozen state and introduce a small set of trainable components, thereby significantly reducing memory and compute overhead~\citep{lin2024awq}. Among them, Low-Rank Adaptation (LoRA)~\citep{lora,liu2024dora,song2024low,buyukakyuz2024olora,zhao2024galore} is a popular and widely adopted approach due to its simplicity, strong empirical performance, and compatibility with modern architectures.

Instead of updating pre-trained model weight directly, LoRA  introduces two learnable low-rank matrices for it, and approximate weight updates through their product. Since the numbers of parameters of these low-rank matrices are much smaller than that of the original pre-trained weights, LoRA significantly reduces the memory overhead during fine-tuning.

Despite its widespread success, LoRA has inherent limitations, particularly in its ability to model complex weight adaptation behaviors. LoRA approximates the weight change with the product of two low-rank matrices. While recent studies have observed that the cumulative weight updates during fine-tuning often exhibit approximately low-rank structure~\citep{zhao2024galore}, LoRA itself learns these updates from scratch using randomly initialized parameters, without leveraging any prior knowledge from the pre-trained weights. As a result, the optimization process becomes more challenging, especially under low-rank settings where the parameter space is highly constrained and prone to suboptimal local minima~\citep{pan2024lisa}. Furthermore, due to its linear structure, LoRA may struggle to capture intricate adaptation patterns required by many downstream tasks. To compensate for this limited capacity, LoRA-based methods often resort to increasing the rank of the update matrices, which in turn reduces their parameter efficiency and undermines their original motivation.

To overcome these limitations, we propose a \textbf{p}arameter-\textbf{e}fficient \textbf{a}daptation method with weight-aware \textbf{n}e\textbf{u}ral \textbf{t}weakers, {\name}, which incorporates a lightweight neural network, which takes the \textit{pre-traiend weight} as the input, into the adaptation process.
Unlike LoRA, which approximates weight updates linearly through low-rank decomposition, {\name} models cumulative weight updates as explicit functions of the pre-trained model’s original weights. This enables {\name} to capture complex, non-linear patterns in the weight space, improving adaptation performance without increasing the number of parameters. The key innovation in {\name} lies in introducing compact neural networks, \textit{neural tweakers}, that transforms the pre-trained weights, approximating the updates with minimal additional computation. This nonlinear transformation enhances the expressiveness of the parameter updates while maintaining the efficiency. Importantly, this architecture facilitates a more efficient exploration of the optimization landscape, leading to better task adaptation, particularly in cases where linear methods like LoRA would require much larger ranks to achieve competitive results. We theoretically demonstrate that {\name} can achieve the same or greater expressivity than LoRA with fewer parameters.

The contributions are summarized as follows: 
\begin{itemize}[leftmargin=1em]
    \item We propose {\name}, a new PEFT method that introduces weight-aware neural tweakers to generate adaptive update signals. The method enables efficient and flexible adaptation beyond linear constraints. To the best of our knowledge, this is the first work to introduce nonlinear adaptation for LoRA-based PEFT methods.
    \item The proposed {\name} enhances model performance while maintaining the efficiency. We theoretically show that {\name} can achieve a possibly improved parameter efficiency compared to LoRA.
    \item We conduct extensive experiments on four benchmarks covering over twenty datasets. The experiments show that the proposed {\name} can outperform baselines on both vision and text tasks.
\end{itemize}
\section{Related Work}
In this section, we provide a concise overview of related work on Parameter-Efficient Fine-Tuning (PEFT) methods. PEFT methods aim to reduce the memory overhead of fine-tuning pre-trained models, enabling fine-tuning in resource-constrained environments. According to \citet{han2024parameter}, PEFT methods can be categorized into: 1)~\textbf{Additive PEFT methods}~\citep{chronopoulou2023adaptersoup,edalati2022krona,lester2021power,wang2024universality,liu2022few}, 2)~\textbf{Selective PEFT methods}~\citep{guo2020parameter,das2023unified,sung2021training,ansell2021composable,zaken2021bitfit,vucetic2022efficient,chen2024large,miao2025coeff,chen2025sparse}, 3)~\textbf{Reparameterized PEFT methods}~\citep{hu2021lora,valipour2022dylora,zhang2023adalora,karimi2021compacter,liu2024dora,kopiczko2023vera}, and 4)~\textbf{Hybrid PEFT methods}~\citep{mao2021unipelt,chen2023parameter,he2021towards,zhang2022neural,zhou2024autopeft}. \textit{Additive PEFT methods}~\citep{chronopoulou2023adaptersoup,edalati2022krona,lester2021power,wang2024universality,liu2022few} introduces a small set of additional trainable parameters strategically placed within the model. One of the most prominent additive PEFT approaches is Adapter~\citep{chronopoulou2023adaptersoup,edalati2022krona,zhao2022tiny}, which involves inserting small adapter layers between pre-trained weight blocks. Prompt Tuning~\citep{wang2024universality,lester2021power,vu2021spot,li2021prefix} is another technique, where learnable vectors, or "soft prompts," are prepended to the input sequence without modifying the model's weights. This method is particularly effective for large-scale models and has inspired variants such as Prefix Tuning~\citep{li2021prefix}. \textit{Selective PEFT} focuses on optimizing the fine-tuning process by selectively adjusting a subset of the model’s parameters rather than introducing additional ones. For instance, Diff Pruning~\citep{guo2020parameter} uses a learnable binary mask to select parameters for fine-tuning. Similarly, FishMask~\citep{sung2021training} and Fish-Dip~\citep{das2023unified} leverage Fisher information to determine parameter importance and identify the most crucial ones for updates. Additionally, BitFit~\citep{zaken2021bitfit} fine-tunes only the bias terms in the model, significantly reducing the number of trainable parameters. \textit{Hybrid PEFT} methods aim to combine the strengths of various existing PEFT techniques to enhance model performance across diverse tasks. UniPELT~\citep{mao2021unipelt} integrates LoRA, prefix-tuning, and adapters within each Transformer block, employing a gating mechanism to determine which module should be active during fine-tuning. S4~\citep{chen2023parameter} further explores the design space by partitioning layers into groups and assigning different PEFT methods to each group. Additionally, NOAH~\citep{zhang2022neural} and AUTOPEFT~\citep{zhou2024autopeft} leverage neural architecture search (NAS) to automatically discover optimal combinations of PEFT techniques tailored to specific tasks. 

\textit{Reparameterized PEFT} methods are most close to our proposed method. Low-Rank Adaptation (LoRA)-based methods, which are representative of reparameterized PEFT approaches, have gained significant attention due to their minimal architectural changes, no additional inference costs, and high efficiency. LoRA~\citep{hu2021lora} introduces two trainable low-rank matrices for each pre-trained model weight to approximate the desired updates of the original model. Extensions of LoRA include DyLoRA~\citep{valipour2022dylora}, which dynamically adjusts the rank of the low-rank matrices during training to optimize for specific tasks; AdaLoRA~\citep{zhang2023adalora}, which adaptively allocates the parameter budget among weight matrices based on their importance scores; and DoRA~\citep{liu2024dora}, which decomposes the pre-trained weight into magnitude and direction, applying LoRA only for direction updates. Other variants include VeRA~\citep{kopiczko2023vera}, which introduces shared frozen random matrices across layers to improve efficiency further, and RoseLoRA~\citep{wang2024roselora}, which employs a row- and column-wise sparse low-rank adaptation mechanism to selectively update the most significant parameters. FourierFT~\citep{gaoparameter} replaces the matrix multiplication in LoRA with a Fourier transform, while PiSSA~\citep{pissa} and MiLoRA~\citep{milora} update the principal and minor singular components of the weight matrix, respectively. However, existing PEFT methods rely on linear transformations to approximate pre-trained weight updates, which struggle to capture the complex relationships inherent in weight updates, leading to a significant performance gap compared to full fine-tuning. Meanwhile, existing research like \citep{teney2024neuralredshift} also demonstrates that nonlinear activation is an integral part of the neural network driving its success.

\begin{figure*}
    \centering
    \includegraphics[width=0.9\textwidth]{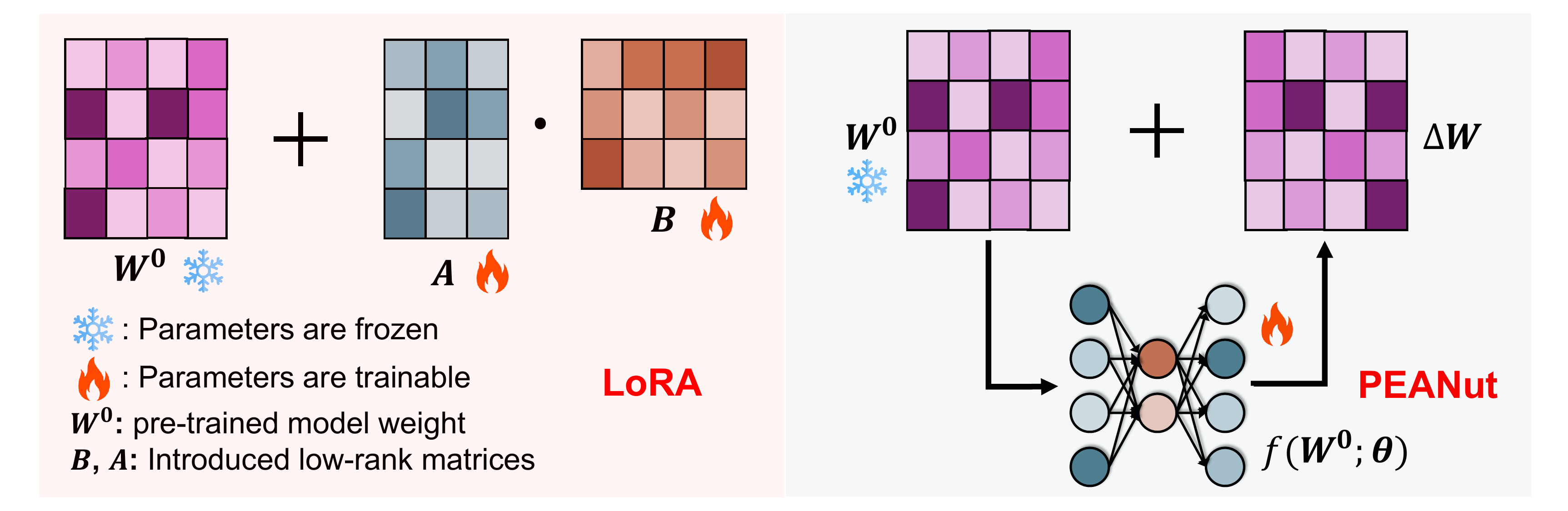}
    \caption{\centering Framework of proposed {\name}.}
    \label{fig:framework}
\end{figure*}

\section{Methodology}
In this section, we start with a brief introduction of LoRA. 
Motivated by a key limitation in LoRA parameter efficiency 
that roots from LoRA parameterization form, we propose {\name}, a novel PEFT method to solve the issue. Notably, {\name} is able to achieves better parameter efficiency provably. 


\subsection{Preliminary}
LoRA~\citep{hu2021lora} assumes that the updates to model weights during the fine-tuning exhibit low-rank properties. 
Built upon this, 
LoRA models the \textit{incremental update} of some weight matrix $\mathbf{W}^{0}\in \mathbb{R}^{d_1 \times d_2}$ in a pre-trained model approximately by the product of two learnable low-rank matrices
\begin{align*}
    \mathbf{W}=\mathbf{W}^{0}+\Delta \mathbf{W}=\mathbf{W}^{0}+\mathbf{AB},
\end{align*}
where $\mathbf{A} \in \mathbb{R}^{d_1 \times r}$ and $\mathbf{B} \in \mathbb{R}^{r \times d_2}$ with $r \ll \min(d_1,d_2)$. 
When conducting fine-tuning, only introduced two low-rank matrices $\mathbf{A}$ and $\mathbf{B}$ will be updated and the pre-trained weight $\mathbf{W}^{0}$ is frozen, as represented by the following optimization
\begin{align}
    \min\nolimits_{\mathbf{A},\mathbf{B}} \ \mathcal{L}(\mathcal{D}_{\text{train}};\mathbf{W}^{0}+\mathbf{AB}),
\label{eq:eq2}
\end{align}
where $\mathcal{D}_{\text{train}}$ is the training set used for fine-tuning and $\mathcal{L}$ is the loss function. 
Since $\mathbf{A}$ and $\mathbf{B}$ are both low-rank matrices that contain significantly fewer parameters compared with the original $\mathbf{W}^{0}$, the LoRA costs much less memory space compared to the fully fine-tuning.

\subsection{Inherent Limitation of LoRA Formulation}
While LoRA family have demonstrated remarkable parameter efficiency in fine-tuning pre-trained models for diverse downstream tasks, 
we argue that their product-based formulation are suboptimal for capturing the full fine-tuning dynamics in an efficient way. 

Specifically, 
when fully fine-tuning a pre-trained model, the update process of weight $\mathbf W$ is typically performed through an iterative gradient descent:
\begin{align*}
    \mathbf{W}^{0}_{t}=\mathbf{W}^{0}_{t-1}-\eta \nabla_{\mathbf{W}^{0}_{t-1}}\mathcal{L},
\end{align*}
where $\mathbf{W}^{0}_{0}=\mathbf{W}^{0}$ is the initial state, $\eta$ is the learning rate, and $\mathbf{W}^{0}_{t}$ represents the weights after $t$ iterations. 
The cumulative change in the weights over time can be represented as:
\begin{align*}
    \Delta \mathbf{W}=\mathbf{W}^{0}_{t}-\mathbf{W}^{0}_{0}.
\end{align*}
This weight change $\Delta \mathbf{W}$ can be interpreted as a function of the original pre-trained weights $\mathbf{W}^{0}$, capturing the model's adaptation to the specific task during fine-tuning. 

Nonetheless, LoRA matrices $\mathbf A$ and $\mathbf B$ are parameterized in a free way without any dependency on $\mathbf{W}^{0}$. 
While gradient $\nabla_{\mathbf A} \mathcal{L}$ and $\nabla_{\mathbf B} \mathcal{L}$ are implicit functions of $\mathbf{W}^{0}$, 
making final learned $\mathbf A_t, \mathbf B_t$ indirectly depends on $\mathbf W^0$ as well, 
as will be proved shortly, 
the lack of explicit dependency still makes LoRA \textit{inherently suboptimal} for fine-tuning pre-trained models.

\subsection{Parameter-Efficient Adaptation with Weight-aware Neural Tweakers}\label{sec: nonlinear}

Motivated by the above analysis on LoRA's limitation, 
we propose to approximate $\Delta \mathbf{W}$ using a lightweight neural network that \textit{explicitly} takes pre-trained model weight $\mathbf{W}^{0}$ as input and outputs the weight update directly. 
By doing so, our approach captures more complex and richer transformation of the weights in a more efficient manner. 
We refer to our method as \textbf{p}arameter-\textbf{e}fficient \textbf{a}daptation method with weight-aware \textbf{n}e\textbf{u}ral \textbf{t}weakers~({\name}).

Following LoRA's updates paradigm,
the proposed {\name} also provides incremental update of pre-trained models.
However, {\name} modifies the forward pass of the model by introducing a dynamic \textit{nonlinear} weight transformation. 
Specifically, the modified model's forward propagation is formulated as:
\begin{align*}
    \bm{y}=(\mathbf{W}^{0}+f(\mathbf{W}^{0};\bm{\theta}))\bm{x}.
\end{align*}
Here $\bm{x}$ and $\bm{y}$ are the input and output with respect to the current layer, respectively, and $f(\cdot;\bm{\theta}): \mathbb{R}^{d_{1}\times d_{2}}\to \mathbb{R}^{d_{1}\times d_{2}}$ is a nonlinear neural network parameterized by learnable parameter $\bm{\theta}$. The neural network $f(\mathbf{W}^{0};\bm{\theta})$ generates the weight update as a function of $\mathbf{W}^{0}$. 

To ensure the parameter efficiency of our {\name}, the learnable neural network $f(\mathbf{W}^{0};\bm{\theta})$ should be lightweight, i.e., the number of parameters $\bm{\theta}$ should be much fewer than that of the original pre-trained weight $\mathbf{W}^{0}$. 
Therefore, we parametrize $f(\mathbf{W}^{0};\bm{\theta})$ as a neural network with \textit{bottleneck} layers. 
For example, a simple case is $f(\mathbf{W}^{0};\bm{\theta})=\sigma(\mathbf{W}^{0}\bm{\Theta}_{1})\bm{\Theta}_{2}$, where $\bm{\theta} = (\bm{\Theta}_1, \bm{\Theta}_2) \in \mathbb{R}^{d_{2}\times r} \times \mathbb{R}^{r \times d_2}$ with $r \ll \min(d_1, d_2)$, 
and $\sigma(\cdot)$ is some non-linear activation function such as ReLU~\citep{glorot2011deep}.
We can also increase the layers or add activation function for the output of $f(\mathbf{W}^{0};\bm{\theta})$ to enhance the model expressiveness. 

During fine-tuning, the optimization objective is to minimize the task-specific loss function, which can be represented as
\begin{align*}
    \min\nolimits_{\bm{\theta}} \ \mathcal{L}(\mathcal{D}_{\text{train}};\mathbf{W}^{0}+f(\mathbf{W}^{0};\bm{\theta})),
\end{align*}
where the original pre-trained weight $\mathbf{W}^{0}$ is frozen, and only neural network parameters $\theta$ are updated.
The overview of {\name} is shown in Fig.~\ref{fig:framework}.

\begin{remark}
The benefit of our new formulation lies in two folds. 
First, our incremental update $f(\mathbf{W}^{0};\bm{\theta})$ is an explicit function of $\mathbf{W}^{0}$, allowing it to capture updates in a more effective way. 
Second, the neural network-based $f(\mathbf{W}^{0};\bm{\theta})$ allows for dynamic, non-linear weight updates that can capture more complex interactions. 
These two advantages make {\name} a more effective and efficient PEFT method than existing LoRA-based approaches.     
\end{remark}


\subsection{Theoretical Analysis}\label{sec:theore}
In this section, we show the theoretical analysis of the sub-optimality of LoRA in terms of parameter efficiency. We prove that {\name} can achieve equivalent or even superior efficiency under certain conditions. Specifically, suppose {\name} adopts the following lightweight architecture, as described in Section~\ref{sec: nonlinear}:
\begin{align*}
f(\mathbf{W}^{0}; \bm{\theta}) = \sigma(\mathbf{W}^{0} \bm{\Theta}_1) \bm{\Theta}_2.
\end{align*}
The following proposition demonstrates that {\name} can match the expressivity of LoRA using fewer parameters under specific conditions. Here, expressivity is measured by the minimum attainable loss.

\begin{proposition}\label{prop: equivalence in loss}
Given pre-trained weight matrix $\mathbf{W}^{0}$. 
Let $\sigma$ denote ReLU activation function, 
and  $\bm{U}^0 \in \mathbb{R}^{d_1 \times \operatorname{rank}(\mathbf{W}^0)}$ be the left singular vectors of $\mathbf{W}^{0}$.
Suppose that the fine-tuning loss $\mathcal L$ is invariant under the the projection of the weight matrix to the left singular space of $\mathbf{W}^0$, i.e.,
$\mathcal{L}(\mathcal{D}_{\text{train}}; \mathbf{W}) = \mathcal{L}(\mathcal{D}_{\text{train}}; \bm{U}^0 \bm{U}^{0 \top} \mathbf{W})$ for any $\mathbf{W} \in \mathbb{R}^{d_1 \times d_2}$.
Then, for any $r \geq 1$,
\begin{align*}
    &\min_{\substack{\bm{\Theta}_1 \in \mathbb{R}^{d_2 \times 2r},\\\bm{\Theta}_2 \in \mathbb{R}^{2r \times d_2}}} \mathcal{L}(\mathcal{D}_{\text{train}}; \mathbf{W}^0 + f(\mathbf{W}^0; (\bm{\Theta}_1, \bm{\Theta}_2))) \\&\leq \min_{\substack{\mathbf{A} \in \mathbb{R}^{d_1 \times r},\\\mathbf{B} \in \mathbb{R}^{r \times d_2}}} \mathcal{L}(\mathcal{D}_{\text{train}}; \mathbf{W}^0 + \mathbf{A} \mathbf{B})\\
    &\leq \min_{\substack{\bm{\Theta}_1 \in \mathbb{R}^{d_2 \times r},\\\bm{\Theta}_2 \in \mathbb{R}^{r \times d_2}}} \mathcal{L}(\mathcal{D}_{\text{train}}; \mathbf{W}^0 + f(\mathbf{W}^0; (\bm{\Theta}_1, \bm{\Theta}_2))).
\end{align*}
\end{proposition}

In words, 
Prop \ref{prop: equivalence in loss} demonstrates the (approximate) equivalence of LoRA and {\name} in terms of their expressivity. Specifically, the minimum attainable loss using rank-$r$ LoRA can be achieved by {\name} with $2r$ hidden units, and conversely, the minimum attainable loss using {\name} with $r$ hidden units can be achieved rank-$r$ LoRA, provided the invariance assumption holds.
This equivalence further implies that the function classes realized by {\name} with $O(r)$ hidden dimensions and rank-$r$ LoRA are equivalent in expressivity, as the result holds for any loss functions. 

Importantly, \textit{this highlights a potential improvement in parameter efficiency by {\name}}. 
Namely, {\name} with $O(r d_2)$ parameters maintains the expressivity of LoRA with $r(d_1+d_2)$ parameters. 
That it to say, {\name} offers a significant improvement in parameter efficiency when $d_2 \ll d_1$~(
a condition that widely holds for the down projection matrix of transformers fully-connected layers~\citep{vaswani2017attention,dosovitskiy2021an}
). 
In such cases, 
{\name} provably achieves better parameter efficiency than LoRA. 
The added parameter efficiency can also improve sample efficiency by allowing the model to learn representations with the same or fewer data points.

The invariance assumption in Proposition \ref{prop: equivalence in loss} pertains to the pre-trained model, and asserts that the later layers of the model depends solely on the task-relevant feature space. Given that we fine-tune a pre-trained model, the later layers are expected to capture this task-relevant feature space, which is described by the left singular space of $\mathbf{W}^0$. In practice, since the later layers primarily rely on this pre-trained feature space, the principal directions of the pre-trained weight matrix, represented by its singular vectors, encode most of the useful features for downstream tasks. This makes the loss largely invariant to changes outside this subspace. The proof is available in Appendix \ref{sec: proof of equivalence}.

If we consider a sinusoid activation function $\sigma_\textnormal{p}(x) = \sin(2\pi x)$, then stronger result that \textbf{{\name} has expressivity (almost) greater than or equal to a LoRA with possibly more parameters can be established without the invariance assumption}. We defer the result to the Appendix \ref{sec: another theory}.

\section{Complexity Analysis}

In this section, we compare the computational and space complexity of {\name} and LoRA. 

\noindent\textbf{Space Complexity.}
Because we set the introduced parameters of LoRA and {\name} to be the same, we only discuss the space complexity of the training in this section. Both LoRA and {\name} require storing the added parameters and their gradients. {\name} may incur a slightly higher activation memory during backpropagation due to the extra nonlinearity, but our empirical results (see Sec.~\ref{sec:runtime}) show that this overhead is minimal and does not affect scalability in practice.

\noindent\textbf{Computational complexity.}
In terms of per-step computation cost, LoRA computes the residual update as $\mathbf{A} \mathbf{B}x$, which costs $\mathcal{O}(d_1 r + r d_2)$ per input vector $x$. {\name} requires computing $f(\mathbf{W}_0; \theta)x$. When using the aforementioned example with one-hidden layer and having the same latent dimension as LoRA, the main cost is $\mathcal{O}(d_1 d_2 r)$. Although this complexity is higher than LoRA's, both methods benefit from highly matrix-friendly implementations. In practice, our experiments (see Sec.~\ref{sec:runtime}) show that the empirical training time per step is comparable. Importantly, during inference, both {\name} and LoRA allow their update modules to be merged into the original weight matrix $\mathbf{W}_0$, ensuring that no additional forward-pass cost is incurred in deployment.
\section{Experiment}
In the experiments, we evaluate the proposed {\name} and answer the following questions: 
\begin{enumerate}[leftmargin=3em]
    \item[\textbf{RQ1}] How does {\name} compare to state-of-the-art PEFT methods on NLP and vision tasks?
    \item[\textbf{RQ2}] What is the role of nonlinear approximation in the proposed {\name}?
    \item[\textbf{RQ3}] What is the real runtime and memory consumption of proposed {\name}?
    \item[\textbf{RQ4}] How does the performance of {\name} vary with different fine-tuned modules, depths of the lightweight neural network, or non-linear activation functions?
\end{enumerate}

\subsection{Benchmarks and Experiment Setups}
We experiment {\name} on datasets from four representative benchmarks: 1)~\textbf{Commonsense Reasoning} covers diverse multi-choice problems from BoolQ~\citep{boolq}, PIQA~\citep{piqa}, SIQA~\citep{siqa}, HellaSwag~\citep{hellaswag}, WinoGrande~\citep{winogrande}, ARC-e and ARC-c~\citep{ARC}, and OpenBookQA~\citep{openbookqa} datasets. Following \citet{milora}, we finetune LLaMA2-7B~\citep{llama2-7b}, LLaMA3-8B \citep{llama3} and Qwen3-8B~\citep{qwen3} on Commonsense170K~\citep{llmadapter} benchmark which combines all previous training sets, and evaluate the accuracy on their testing sets separately. 2)~\textbf{Arithmetic Understanding} consists of two math reasoning datasets: GSM8K~\citep{gsm8k} and MATH~\citep{MATH}. We finetune LLaMA2-7B~\citep{llama2-7b} and Qwen3-8B~\citep{qwen3} on MetaMath~\citep{metamath} dataset following \citet{milora}. 
Models need to generate correct answers, and accuracy is used as the evaluation metric. 3)~\textbf{Natural Language Understanding} consists of eight datasets from the GLUE benchmark~\citep{glue}. We follow the evaluation metrics and setups from \citet{fourierft,wuandarora2024reft}. 4)~\textbf{Image Classification} consists of Oxford-Pets~\citep{pets}, CIFAR10~\citep{cifar}, DTD~\citep{dtd}, EuroSAT~\citep{euro}, RESISC45~\citep{resisc}, StanfordCars~\citep{cars}, FGVC~\citep{fgvc}, and CIFAR100~\citep{cifar} following \citet{fourierft}. The first five datasets have small label spaces, while the last three have large label spaces.

\textbf{Baselines methods} are constructed on a task basis. Specifically, for each task, the proposed {\name} is compared with representative baselines from the corresponding domain. For both Commonsense Reasoning and Arithmetic Understanding, following \citet{milora}, LoRA~\citep{lora}, PiSSA~\citep{pissa} and MiLoRA~\citep{milora} are employed as baselines. {\name} is applied to \texttt{query, key, value, MLP up} and \texttt{MLP down} layers. For Natural Language Understanding, we follow the setup from prior works~\citep{fourierft,wuandarora2024reft} that evaluate various representative PEFT methods, including LoRA~\citep{lora}, Adapter~\cite{houlsby2019parameter}, BitFit~\citep{zaken2021bitfit}, RED~\citep{wu2024advancing}, DoRA~\citep{liu2024dora}, ReFT~\cite{wuandarora2024reft}, and FourierFT~\citep{fourierft}. For Image Classification, we follow the setting of \citet{fourierft} and take linear probing (LP), LoRA~\citep{lora} and FourierFT~\citep{fourierft} as baselines. {\name} is applied to the query and value layers. See our appendix for details about the datasets (App \ref{datasets}) and hyper-parameters (App \ref{hyper}).

\subsection{Performance Comparison}

We showcase {\name} performance on different tasks. 

\noindent\textbf{Commonsense Reasoning.}
We experiment {\name} with eight commonsense reasoning datasets to address RQ1, 
results are shown in Tab~\ref{tab:commonsense}. 
We compare the performance of three state-of-the-art baselines with the proposed {\name}, 
and {\name} consistently outperforms all of them, achieving the highest accuracy on all tasks. 
Specifically, {\name} surpasses LoRA, PiSSA, and MiLoRA in terms of average accuracy by 4.6\%, 10\%, and 2.5\%, respectively, when using LLaMA2-7B as the backbone. On LLaMA3-8B as the backbone, {\name} demonstrates average improvements of 4.9\%, 11.8\%, and 2.9\% over LoRA, PiSSA, and MiLoRA, respectively. With Qwen3-8B as the backbone, {\name} improves average accuracy over LoRA and MiLoRA by 3.9\% and 2.4\%, respectively. These results highlight the effectiveness and superiority of {\name} as a PEFT method.

\noindent\textbf{Arithmetic Reasoning.}
In this section, we present results on two arithmetic reasoning tasks in Tab~\ref{tab:math} to help address RQ1. 
From the table, 
while full fine-tuning~(FFT) achieves highest accuracy across the two datasets, 
the performance gap between the proposed {\name} and FFT is very small, 
despite that {\name} relies on significantly fewer trainable parameters. 
Moreover, compared to state-of-the-art PEFT baselines, {\name} achieves remarkable performance improvements. 
In terms of average accuracy, {\name} demonstrates improvements of 7.5\%, 12.4\%, and 2.4\% over LoRA, PiSSA, and MiLoRA, respectively, when using LLaMA2-7B as the backbone. With Qwen3-8B as the backbone, {\name} improves average accuracy over LoRA and MiLoRA by 7.1\% and 3.7\%. These results on clearly confirm that {\name} is highly effective and efficient for complex reasoning tasks.

\noindent\textbf{Natural Language Understanding.} We further conduct experiments on the GLUE to answer RQ1, results are shown in Tab~\ref{tab:glue}. From the table, {\name} significantly outperforms state-of-the-art PEFT methods. Specifically, {\name}-S, which uses a similar number of trainable parameters as FourierFT~\citep{fourierft}, DiReFT~\citep{wuandarora2024reft}, and LoReFT~\citep{wuandarora2024reft}, surpasses all PEFT baselines and experiences only a small performance drop (0.2\%) compared to FFT. Additionally, {\name}-L exceeds the performance of all baselines, including FFT, with roughly the same number of trainable parameters as in LoRA. These results demonstrate that {\name} exhibits excellent generalization ability while maintaining great parameter efficiency.

\noindent\textbf{Image Classification.} In this section, we conduct experiments on image classification tasks to address RQ2, 
{\name} uses depth of 6, and 
results are shown in Tab~\ref{tab:cv}. 
From the table, {\name} significantly outperforms LoRA and FourierFT using the same number of trainable parameters. Specifically, {\name} achieves performance improvements of 11.05\%, 7.30\%, and 26.02\% compared to LoRA, FourierFT, and LP, respectively. Furthermore, compared to FFT, the proposed {\name} shows negligible performance drop (86.49\% v.s. 86.34\%), while using only 0.3\% of the trainable parameters required by FFT. This demonstrates that {\name} exhibits exceptional adaptation capability not only on NLP tasks, but also on vision tasks as well. Additionally, it verifies the effectiveness of the nonlinear adaptation used in {\name}.
\begin{table*}[ht!]
    \centering
    \caption{
    Common Reasoning performance of {\name} and PEFT baselines on LLaMA 2-7B, LLaMA 3-8B and Qwen 3-8B. Results marked with ``{+}'' are taken from \citet{liu2024dora}, and those marked with ``$\ast$'' are taken from \citet{milora}. Best results are in \textbf{bold}. ``AVG'' means the average accuracy of all datasets.}
    \adjustbox{max width=\textwidth}{
    \begin{tabular}{ccccccccccc}
\toprule
        \multirow{2}{*}{\textbf{Model}} & \multirow{2}{*}{\textbf{PEFT}} & \multicolumn{8}{c}{\textbf{Accuracy} ($\uparrow$)} \\
        \cmidrule{3-11}
        & & \textbf{BoolQ} & \textbf{PIQA} & \textbf{SIQA} & \textbf{HellaSwag} & \textbf{WinoGrande} & \textbf{ARC-e} & \textbf{ARC-c} & \textbf{OBQA} & \textbf{AVG} \\
        \midrule
        \multirow{4}{*}{LLaMA2-7B} &  LoRA\textsuperscript{+} &  69.8 & 79.9 & 79.5 & 83.6 & 82.6 & 79.8 & 64.7 & 81.0 & 77.6 \\ 
         & PiSSA\textsuperscript{*} & 67.6 & 78.1 & 78.4 & 76.6 & 78.0 & 75.8 & 60.2 & 75.6 & 73.8 \\
         & MiLoRA\textsuperscript{*} & 67.6 & 83.8 & 80.1 & 88.2 & 82.0 & 82.8 & 68.8 & 80.6 & 79.2 \\\cmidrule{2-11}
         & {\name} & \textbf{71.9} & \textbf{84.0} & \textbf{80.4} & \textbf{88.9} & \textbf{84.6} & \textbf{86.5} & \textbf{71.6} & \textbf{83.0} & \textbf{81.4} \\
         \midrule
       \multirow{4}{*}{LLaMA3-8B} &  LoRA\textsuperscript{+} &  70.8 & 85.2 & 79.9 & 91.7 & 84.3 & 84.2 & 71.2 & 79.0 & 80.8 \\
         & PiSSA\textsuperscript{*} & 67.1 & 81.1 & 77.2 & 83.6 & 78.9 & 77.7 & 63.2 & 74.6 & 75.4 \\
         & MiLoRA\textsuperscript{*} & 68.8 & 86.7 & 77.2 & 92.9 & 85.6 & 86.8 & 75.5 & 81.8 & 81.9 \\ \cmidrule{2-11}
         & {\name} & \textbf{72.1} & \textbf{87.0} & \textbf{80.9} & \textbf{94.3} & \textbf{86.7} & \textbf{91.4} & \textbf{78.9} & \textbf{84.8} & \textbf{84.5} \\
         \midrule
         \multirow{3}{*}{Qwen3-8B} 
         & LoRA & 86.3 & 87.2 & 84.1 & 92.5 & 81.5 & 89.6 & 78.8 & 89.5 & 86.2 \\
         & MiLoRA & 85.2 & 89.3 & 84.2 & 94.6 & 82.2 & 92.3 & 82.7 & 89.5 & 87.5 \\ 
         \cmidrule{2-11}
         & {\name} & \textbf{89.4} & \textbf{90.2} & \textbf{87.4} & \textbf{95.7} & \textbf{85.5} & \textbf{92.7} & \textbf{82.6} & \textbf{93.6} & \textbf{89.6} \\
        \bottomrule\\
        \label{tab:commonsense}
    \end{tabular}
    }
\end{table*}

\renewcommand{\arraystretch}{0.91}
\begin{table*}[t!]
\centering
\caption{
Image Classification performance on ViT-base. 
Best results are in \textbf{bold}. 
``AVG'' means the average accuracy of all datasets. 
Results marked with ``$\ast$'' are taken from \citet{fourierft}.
}
\label{tab:cv}
\addtolength{\tabcolsep}{-3.5pt}
\resizebox{0.95\textwidth}{!}{%
\begin{tabular}{@{}l|c|ccccccccc@{}}
\toprule
\textbf{Method} & \begin{tabular}[c]{@{}c@{}} \textbf{Params} (M)\end{tabular} & \textbf{\small OxfordPets} & \textbf{\small StanfordCars} & \textbf{\small CIFAR10} & \textbf{\small DTD} & \textbf{\small EuroSAT} & \textbf{\small FGVC} & \textbf{\small RESISC45} & \textbf{\small CIFAR100} & \textbf{\small AVG} \\ \midrule
\small FFT$^*$ & 85.8M & 93.14 & 79.78 & \bf98.92 & 77.68 & \bf99.05 & \bf54.84 & \bf96.13 & \bf92.38 & \bf86.49\\ \midrule
\small LP$^*$ & - & 90.28 & 25.76 & 96.41 & 69.77 & 88.72 & 17.44 & 74.22 & 84.28 & 68.36\\
\small LoRA$^*$ & 581K & 93.19 & 45.38 & 98.78 & 74.95 & 98.44 & 25.16 & 92.70 & 92.02 & 77.58\\
\small FourierFT$^*$ & 239K & 93.05 & 56.36 & 98.69 & 77.30 & 98.78 & 32.44 & 94.26 & 91.45 & 80.29\\ 
\midrule
{\name} & 263K & \bf93.62 & \bf80.21 & 98.78 & \bf79.61 & 98.85 & 52.93 & 94.71 & \bf92.02 & 86.34\\ 
\bottomrule
\end{tabular}%
}
\end{table*}

\begin{table*}[ht!]
\renewcommand{\arraystretch}{0.91}
    \centering
    \caption{
    GLUE benchmark performance on RoBERTa-base. Results marked with ``$\ast$'' are taken from \citet{wu2024advancing}.
    Best results are in \textbf{bold}. ``AVG'' means the average accuracy of all datasets. 
    {\name-S} applies trainable modules to layers starting from the 4th layer, with hidden dimensions set to 1. 
    This matches the parameter numbers of FourierFT. 
    {\name-L} applies {\name} to all layers with hidden dimension 8, aligning the parameter budget of LoRA.
    }
    \adjustbox{max width=0.95\textwidth}{
    \begin{tabular}{lcccccccccc}
\toprule
        \multirow{2}{*}{\textbf{PEFT}} & \multirow{2}{*}{\textbf{Params} (\%)} & \multicolumn{8}{c}{\textbf{Accuracy} ($\uparrow$)} \\
        \cmidrule{3-11}
        & & \textbf{MNLI} & \textbf{SST-2} & \textbf{MRPC} & \textbf{CoLA} & \textbf{QNLI} & \textbf{QQP} & \textbf{RTE} & \textbf{STS-B} & \textbf{AVG} \\
        \midrule
        FFT &  100\% &  87.3 & 94.4 & 87.9 & 62.4 & 92.5 & 91.7 & 78.3 & 90.6 & 85.6 \\ \cmidrule{2-11}
        Adapter$^*$ & 0.318\% & 87.0 & 93.3 & 88.4 & 60.9 & 92.5 & \textbf{90.5} & 76.5 & 90.5 & 85.0 \\
        LoRA$^*$ & 0.239\% & 86.6 & 93.9 & 88.7 & 59.7 & \bf92.6 & 90.4 & 75.3 & 90.3 & 84.7 \\
        Adapter\textsuperscript{FNN}$^*$ & 0.239\% & \textbf{87.1} & 93.0 & 88.8 & 58.5 & 92.0 & 90.2 & 77.7 & 90.4 & 84.7 \\
        BitFit$^*$ & 0.080\% & 84.7 & 94.0 & 88.0 & 54.0 & 91.0 & 87.3 & 69.8 & 89.5 & 82.3 \\
        RED$^*$ & 0.016\% & 83.9 & 93.9 & 89.2 & 61.0 & 90.7 & 87.2 & 78.0 & 90.4 & 84.3 \\
        FourierFT & 0.019\% & 84.7 & 94.2 & 90.0 & 63.8 & 92.2 & 88.0 & 79.1 & \bf90.8 & 85.3 \\
        DiReFT$^*$ & 0.015\% & 82.5 & 92.6 & 88.3 & 58.6 & 91.3 & 86.4 & 76.4 & 89.3 & 83.2 \\
        LoReFT$^*$ & 0.015\% & 83.1 & 93.4 & 89.2 & 60.4 & 91.2 & 87.4 & 79.0 & 90.0 & 84.2 \\
        \cmidrule{2-11}
        {\name-S} & 0.019\% & 84.9 & 94.3 & 90.2 & 64.6 & 92.0 & 88.3 & 78.3 & 90.5 & 85.4 \\
        {\name-L} & 0.241\% & 86.9 & \textbf{95.2} & \textbf{90.0} & \bf64.8 & 92.3 & 90.3 & \bf82.7 & 90.7 & \textbf{86.6} \\
        \bottomrule\\
    \end{tabular}
    }
    \label{tab:glue}
\end{table*}

\begin{table}[ht!]
    \centering
    \caption{
    Arithmetic Reasoning performance on LLaMA 2-7B and Qwen 3-8B. Results marked with ``{+}'' are taken from \citet{metamath}, and those marked with ``$\ast$'' are taken from \citet{milora}. Best results are in \textbf{bold}. ``AVG'' means the average accuracy of all datasets.}
    \label{tab:math}
    \label{tab:math}
    \begin{tabular}{cc|ccc}
    \toprule
    \textbf{Model} & \textbf{Method} & \textbf{GSM8K}& \textbf{MATH}& \textbf{AVG} \\ 
    \midrule
    \multirow{5}{*}{LLaMA2-7B}
    & FFT \textsuperscript{+}& 66.50& 19.80& 43.20\\ \cmidrule{2-5}
    & LoRA\textsuperscript{*}& 60.58& 16.88& 38.73\\
    & PiSSA\textsuperscript{*}& 58.23& 15.84&37.04\\
    & MiLoRA\textsuperscript{*}& 63.53& 17.76&40.65\\ 
    \cmidrule{2-5}
    & {\name} & \bf65.05& \bf18.30& \bf41.68\\ 
    \midrule
    \multirow{3}{*}{Qwen3-8B}
    & LoRA & 85.22 & 67.26 & 76.24\\ 
    & MiLoRA & 89.01 & 68.58 & 78.80\\
    \cmidrule{2-5}
    & {\name} & \bf92.87 & \bf70.50 & \bf81.69\\
    \bottomrule
    \end{tabular}%

\end{table}

\subsection{Ablation Study}

In this section, in order to answer RQ2, we present an ablation study with two variants of LoRA to validate the effectiveness of our proposed framework: 
1)~nonlinear LoRA $\bm{y}=(\bm{W_{0}+\sigma(\mathbf{A})\mathbf{B}})\bm{x}$, 
and 2)~multiplicative LoRA $\bm{y}=(\mathbf{W}_{0}+\mathbf{W}_{0}\mathbf{A}\mathbf{B})\bm{x}$. 
Experiments are conducted on image classification benchmarks, and results are reported in Tab~\ref{tab:ablation}. 
According to the table, both nonlinear LoRA and multiplicative LoRA perform worse than {\name}. 
This highlights the effectiveness of incorporating nonlinear approximations and explicitly using model weights as input to the nonlinear function in {\name}.

\renewcommand{\arraystretch}{1.2}
\begin{table*}[t!]
\centering
\caption{
Ablation Study on image classification task. 
The parameters count is the same and ``AVG'' means the average accuracy of all datasets. 
For simple and fair comparison, {\name} uses depth of 2.
}
\label{tab:ablation}
\resizebox{0.8\textwidth}{!}{%
\begin{tabular}{@{}l|ccccccccc@{}}
\toprule
\textbf{Method} & \textbf{\small OxfordPets} & \textbf{\small StanfordCars} & \textbf{\small CIFAR10} & \textbf{\small DTD} & \textbf{\small EuroSAT} & \textbf{\small FGVC} & \textbf{\small RESISC45} & \textbf{\small CIFAR100} & \textbf{\small AVG} \\ \midrule
Nonlinear LoRA & 94.11 & 72.84 & 98.68 & 79.16 & 98.61 & 39.33 & 93.79 & 92.38 & 83.31 \\
Multiplicative LoRA & 93.57 & 77.32 & 98.68 & 77.57 & 98.81 & 46.79 & 94.34 & 91.86 & 84.81 \\
\name & 93.77 & 80.03 & 98.70 & 77.57 & 98.79 & 53.60 & 94.27 & 92.47 & 86.15 \\
\bottomrule
\end{tabular}%
}
\end{table*}

\begin{table*}[ht!]
    \centering
    \caption{Accuracy comparison of {\name} using RoBERTa-base with different depth configurations on the GLUE benchmark. The highest accuracy of methods per category are in \textbf{bold}. ``AVG'' means the average accuracy of all datasets.}
    \vspace{-5pt}
    \adjustbox{max width=0.9\textwidth}{
    \begin{tabular}{lcccccccccc}
\toprule
        \multirow{2}{*}{\textbf{depth}} & \multirow{2}{*}{\textbf{Params} (\%)} & \multicolumn{8}{c}{\textbf{Accuracy} ($\uparrow$)} \\
        \cmidrule{3-11}
        & & \textbf{MNLI} & \textbf{SST-2} & \textbf{MRPC} & \textbf{CoLA} & \textbf{QNLI} & \textbf{QQP} & \textbf{RTE} & \textbf{STS-B} & \textbf{AVG} \\
        \midrule
        2 & 0.239\% & 86.6 & 94.6 & 90.0 & 64.4 & \bf92.7 & 89.7 & 78.7 & \bf90.9 & 86.0 \\
        4 & 0.239\% & 86.7 & 94.5 & \bf90.2 & \bf65.1 & 92.4 & \bf90.5 & 80.5 & 90.8 & 86.3 \\
        6 & 0.241\% & \bf86.9 & \bf95.2 & 90.0 & 64.8 & 92.3 & 90.3 & \bf82.7 & 90.7 & \bf86.6 \\
        \bottomrule\\
    \end{tabular}
    }
    \label{tab:glue_layers}
\end{table*}

\begin{figure}[ht]
    \centering
    \includegraphics[width=0.9\columnwidth]{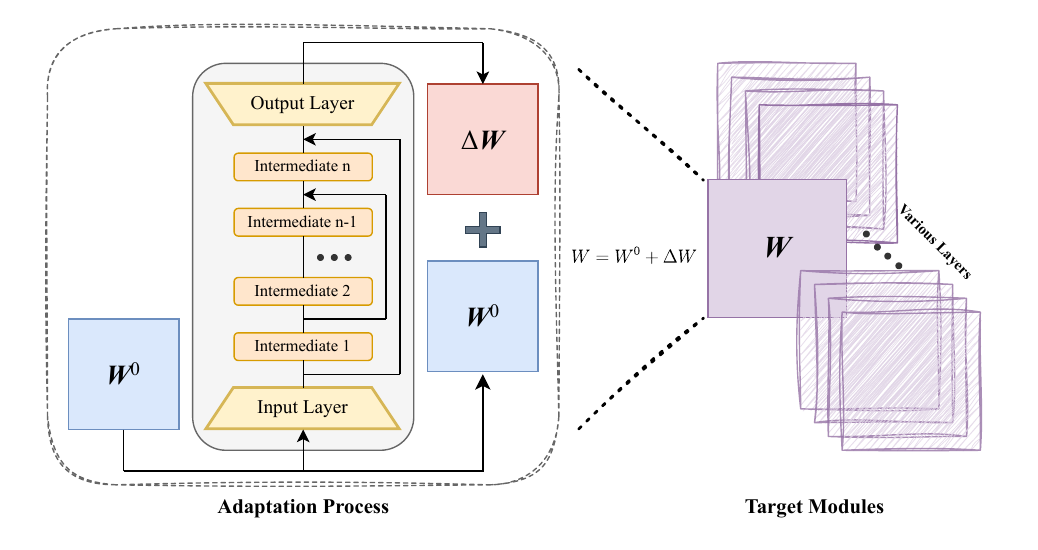}
    \caption{Implementation of introducing more depths to {\name}t. We insert multiple intermediate layers into the layers from vanilla {\name}, with non-linear activation in between. The depth is described as the number of layers in {\name}, with vanilla {\name} having a depth of 2 (i.e. the input and output layers).}
    \label{fig:implementation}
\end{figure}

\begin{table}[h]
\centering
\caption{\centering Runtime and memory consumption of proposed {\name}.}
\label{tab:runtime}
\begin{tabular}{@{}cccc@{}}
\toprule
Dataset                                                                          & Method               & Time   & Memory \\ \midrule
\multirow{2}{*}{MRPC}                                                            & LoRA                 & 77.7s  & 6916MB \\
                                                                                 & \name & 78.4s  & 6916MB \\ \midrule
\multirow{2}{*}{SST-2}                                                           & LoRA                 & 870.7s & 2410MB \\
                                                                                 & \name & 911.4s & 2410MB \\ \midrule
\multirow{2}{*}{\begin{tabular}[c]{@{}c@{}}Commonsense\\ Reasoning\end{tabular}} & LoRA                 & 5.6h   & 22.7GB \\
                                                                                 & \name & 5.7h   & 23.8GB \\ \bottomrule
\end{tabular}%
\end{table}

\subsection{Runtime and Memory Cost}
\label{sec:runtime}
To answer RQ3, we evaluate the computational efficiency of our proposed method, {\name}, by measuring its runtime and memory consumption across three representative datasets: MRPC, SST-2, and Commonsense Reasoning. Table~\ref{tab:runtime} summarizes the results, comparing {\name} against the LoRA approach under identical settings. For a fair comparison, we ensure that the number of trainable parameters is matched between {\name} and LoRA. All experiments are conducted on the same hardware setup using a single NVIDIA A100 GPU. As shown in Table~\ref{tab:runtime}, {\name} exhibits comparable runtime and memory usage to LoRA across all tasks. On the MRPC and SST-2 datasets, {\name} incurs only marginal overhead in training time, with identical memory consumption. For the larger Commonsense Reasoning dataset, {\name} takes 5.7 hours and 23.8GB of memory, compared to LoRA's 5.6 hours and 22.7GB. The slightly higher memory usage is attributed to the additional nonlinear transformation module in {\name}, but the increase remains slight. Overall, the results demonstrate that {\name} achieves improved performance (as discussed in earlier sections) with negligible additional cost in training runtime and memory, highlighting its practicality and scalability for real-world deployment.
\begin{figure}[t]
    \centering
    \includegraphics[width=0.8\textwidth]{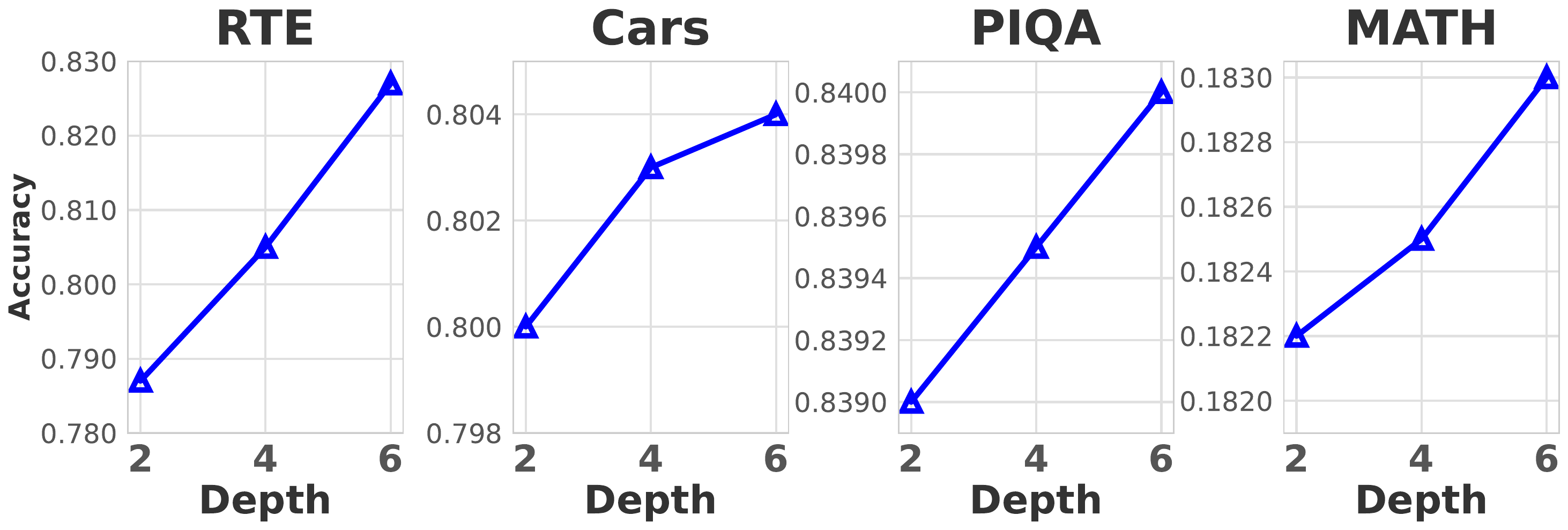}
    \caption{Accuracy on the RTE, StanfordCars, PIQA and MATH  dataset with varying depths of the neural network used in {\name}. The depth here represents the total number of layers in the neural network. We choose depth equals to 2, 4 and 6 layers in the figure.}
    \label{fig:depth}
\end{figure}

\begin{figure*}[t]
    \centering
    \includegraphics[width=0.8\linewidth]{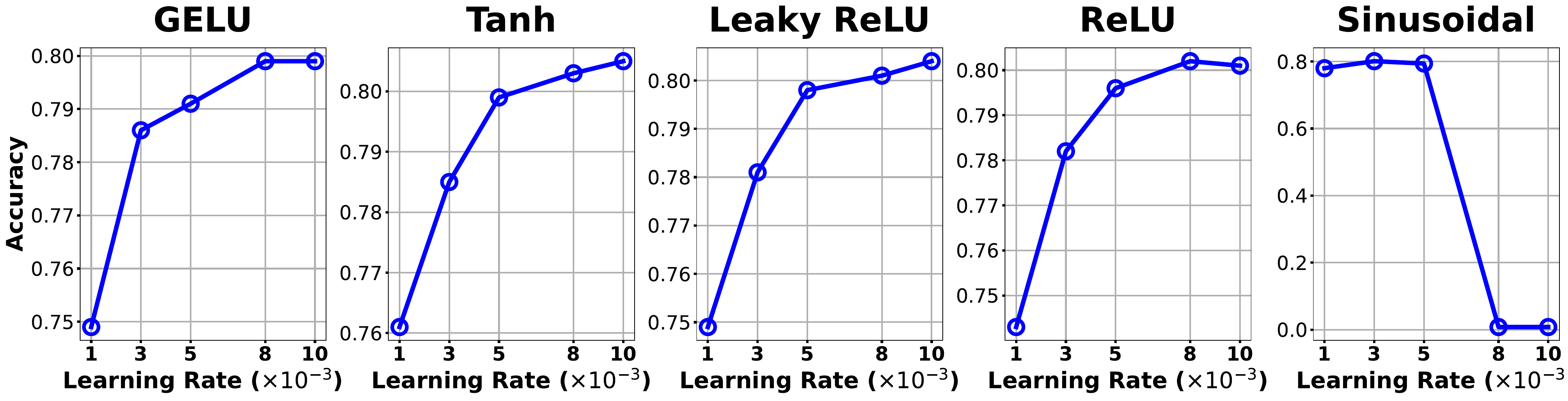}
    \caption{
    Influence of different nonlinear activations choices for {\name}. 
    Experiments are conducted on StanfordCars, {\name} depth is fixed to 2. 
    Different activations share a similar pattern of dependency on learning rate. 
    }
    \label{fig:neat_multiple_nonlinear}
\end{figure*}

\begin{figure*}[h!]
    \centering
    \includegraphics[width=0.8\linewidth]{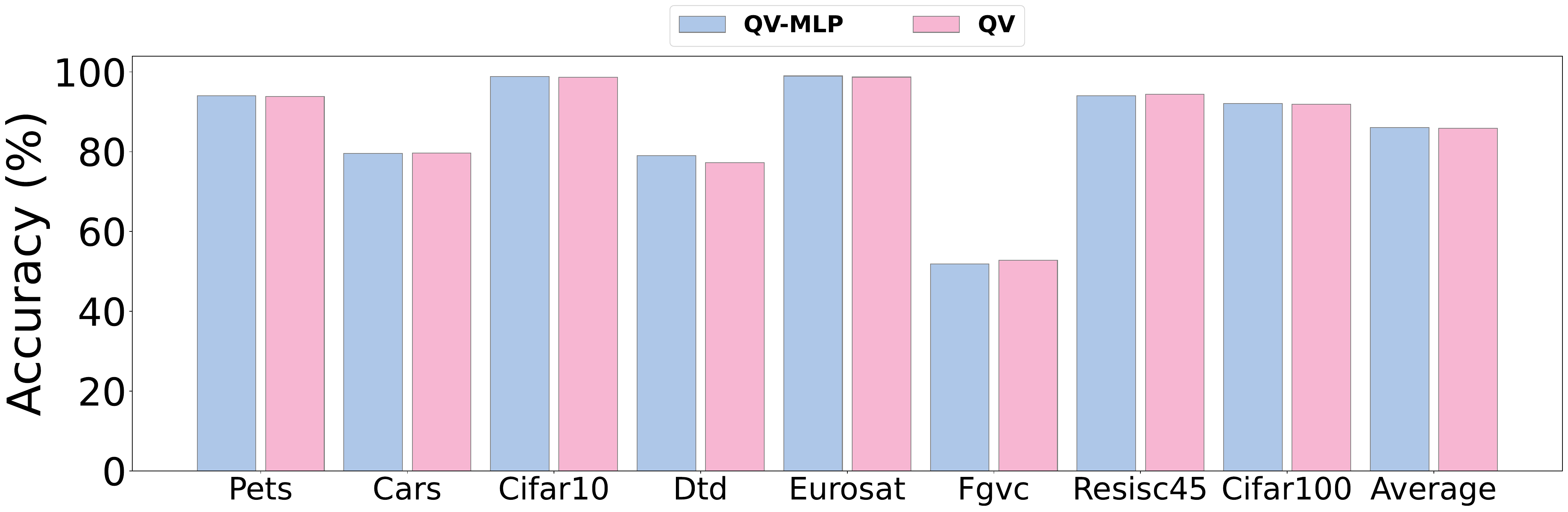}
    \caption{Accuracy of {\name} with different targeted fine-tuning modules, including just QV layers and a combination of QV and MLP layers, on image classification datasets.}
    \label{fig:module-part-analysis}
\end{figure*}

\subsection{Sensitivity w.r.t. Depth}
\label{depth-analysis}

To answer RQ4, we analyze the impact of depth on the performance of {\name}. Deeper architectures are generally more expressive and can better model the complex, nonlinear relationships involved in ideal weight updates \citep{raghu2017expressive}. We evaluate {\name} with varying depth across NLU, vision, commonsense reasoning, and arithmetic reasoning tasks.

We increase the number of intermediate layers inserted between {\name}'s input and output projections. Each intermediate layer is a small feedforward block of shape $\mathbb{R}^{r \times r}$ with non-linear activations. These layers are lightweight compared to the input/output projections ($\mathbf{A} \in \mathbb{R}^{d_2 \times r}$, $\mathbf{B} \in \mathbb{R}^{r \times d_2}$), and add minimal overhead since $r \ll d_2$. The adaptation starts from the frozen base weight $\mathbf{W}^0$, which is transformed through multiple layers to predict $\Delta \mathbf{W}$. We adopt residual connections for stable optimization and improved convergence. All other hyperparameters are kept fixed during this analysis. The layer structure is illustrated in Fig.~\ref{fig:implementation}.

The results in Table~\ref{tab:glue_layers} (GLUE) and Fig.~\ref{fig:depth} (RTE, Cars, PIQA, MATH) indicate that increasing depth consistently improves accuracy. For instance, average GLUE accuracy increases from 86.0 to 86.6 when moving from 2 to 6 layers, with no significant change in parameter count. On other benchmarks, deeper configurations yield steady gains up to 6 layers. Beyond this, performance may slightly drop (e.g., at depth 10), likely due to optimization difficulties without fine-grained hyperparameter tuning.

In summary, depth enhances {\name}'s effectiveness across tasks, offering better adaptation capability with negligible cost in memory or parameters. However, very deep settings may require further tuning to maintain stability.

\subsection{Sensitivity w.r.t. Activations}
\label{various-non-linear}

One key innovation of {\name} compared to LoRA and other PEFT methods, which rely solely on linear transformations for modeling weight updates, is the introduction of non-linear activations within the adaptation neural network. 
Since the choice of non-linear activations directly affects the learning process and the dynamics of weight updates, we investigate how different non-linear activations affects the adaptation performance to address RQ4. 
To this end, we perform experiments on the StanfordCars benchmark using various non-linear activations, including ReLU, Leaky ReLU, GELU, Tanh, and sinusoidal activation ($\sigma_\textnormal{p}(x) = \sin(2\pi x)$). Corresponding results are presented in Fig~\ref{fig:neat_multiple_nonlinear}. 
To ensure a fair comparison, the number of trainable parameters is fixed. 
We optimize other hyperparameters such as learning rate for better performance.

From the figure, the best performance achieved by different activation functions is similar, indicating that the adaptation potential of various activations is comparable. 
This implies that {\name} can benefit from various type of nonlinearity induced by different activations. 
However, it is also worth noting that sinusoidal activations encounters a performance drop at large learning rates. 
Consequently, tuning basic hyperparameters such as learning rate can still be beneficial. 
In conclusion, we suggest ReLU as a default choice in execution, given its practical simplicity~\citep{teney2024neuralredshift}.

\section{Sensitivity w.r.t. Fine-tuned Module}
\label{cv-settings}

We end up this section with a study on applying {\name} to different modules in a ViT, to help better understand RQ4. 

Specifically, 
given the importance of MLP in Transformer architecture, we compare two settings: 1) Following \citet{lora}, we apply {\name} to the query and value layers (QV layers) in the multi-head self-attention module (MHSA) in ViT.
2) Besides QV layers, we also apply {\name} to MLP layers. 
We tune the hidden dimension $r$ to ensure the same parameter scale for fair comparison, and tune the hyperparameters to maximize performance. 
Corresponding results are shown in Fig. \ref{fig:module-part-analysis}.

From the figure, 
applying {\name} to the QV layers yields results comparable to applying {\name} to both the QV and MLP layers. 
This indicates that {\name} is robust to the selections of fine-tuning different modules. This finding confirms another key advantage of {\name}: it does not require extensive manual tuning on which parts (modules, layers) of the foundation model {\name} should be applied. Consequently, {\name} can be easily incorporated to a wide range of scenarios. 
\section{Conclusion}
In this work, we propose {\name}, a novel parameter-efficient fine-tuning~(PEFT) method that introduces nonlinear transformations to enhance model adaptation while maintaining efficiency. By incorporating a lightweight neural network that models cumulative weight updates as functions of the pre-trained weights, {\name} effectively captures complex, nonlinear structures in the weight space, allowing for more expressive and accurate adaptation to downstream tasks. Our theoretical analysis supports the efficacy of {\name}, demonstrating that it can achieve greater or equivalent expressiveness compared to existing LoRA, a popular and state-of-the-art PEFT method, with fewer number of parameters. Through extensive experiments on four benchmarks encompassing over twenty datasets with various pre-trained backbones, {\name} demonstrated superior performance on both NLP and vision tasks compared to existing state-of-the-art methods. 
\clearpage
\bibliography{main}

\appendix
\newpage

\appendix

\section{\centering Appendix}

\renewcommand{\thesection}{\Alph{section}}

\section{Details of Theoretical Results}

In this section, we provide the proof of Proposition \ref{prop: equivalence in loss} and introduce additional theoretical results when we assume sinusoid activation.

\subsection{Proof of Proposition \ref{prop: equivalence in loss}}\label{sec: proof of equivalence}

The intuition behind the proof is that we can always restore an identity function using two ReLU activation functions, i.e., $x = \sigma(x) - \sigma(-x)$ for any $x \in \mathbb{R}$
\begin{proof}[Proof of Proposition \ref{prop: equivalence in loss}]
    We first show that 
    \begin{small}
    \begin{align*}
        &\min_{\bm{\Theta}_1 \in \R^{d_2 \times 2r}, \bm{\Theta}_2 \in \R^{2r \times d_2}} \mathcal{L}(\mathcal{D}_{\text{train}}; \mathbf{W}^0 + f(\mathbf{W}^0; (\bm{\Theta}_1, \bm{\Theta}_2))) \\&\leq \min_{\mathbf{A} \in \R^{d_1 \times r}, \mathbf{B} \in \R^{r \times d_2}} \mathcal{L}(\mathcal{D}_{\text{train}}; \mathbf{W}^0 + \mathbf{A} \mathbf{B}).
    \end{align*}
    \end{small}
    Let $(\mathbf{A}^*, \mathbf{B}^*) = \argmin_{\mathbf{A} \in \R^{d_1 \times r}, \mathbf{B} \in \R^{r \times d_2}} \mathcal{L}(\mathcal{D}_{\text{train}}; \mathbf{W}^0 + \mathbf{A} \mathbf{B})$.
    Take $\bm{\Theta}_1^\# := [(\mathbf{W}^0)^\dag \mathbf{A}^*; - (\mathbf{W}^0)^\dag \mathbf{A}^*] \in \R^{d_2 \times 2r}$ and $\bm{\Theta}_2^\# := [\mathbf{B}^{* \top}; -\mathbf{B}^{* \top}]^\top \in \R^{2r \times d_2}$, where $(\mathbf{W}^0)^\dag \in \R^{d_2 \times d_1}$ is the Moore-Penrose inverse of $\mathbf{W}^0$. Then, since $\sigma$ is a ReLU activation function,
    \begin{small}
    \begin{align*}
        &f(\mathbf{W}^0; (\bm{\Theta}_1^\#, \bm{\Theta}_2^\#)) \\=& \sigma(\mathbf{W}^0 \bm{\Theta}_1^\#) \bm{\Theta}_2^\#\\
        = &\sigma(\mathbf{W}^0 (\mathbf{W}^0)^\dag \mathbf{A}^*) \mathbf{B}^* - \sigma(- \mathbf{W}^0 (\mathbf{W}^0)^\dag \mathbf{A}^*) \mathbf{B}^*\\
        =& \mathbf{W}^0 (\mathbf{W}^0)^\dag \mathbf{A}^* \mathbf{B}^*.
    \end{align*}
    \end{small}
    Note that $\mathbf{W}^0 (\mathbf{W}^0)^\dag = \bm{U}^0 \bm{U}^{0 \top}$ is the projection to the left singular space of $\mathbf{W}^0$. Hence
    \begin{small}
    \begin{align*}
        &\mathcal{L}(\mathcal{D}_{\text{train}}; \mathbf{W}^0 + f(\mathbf{W}^0; (\bm{\Theta}_1^\#, \bm{\Theta}_2^\#))) \\= &\mathcal{L}(\mathcal{D}_{\text{train}}; \bm{U}^{0} \bm{U}^{0 \top} \mathbf{W}^0 + \bm{U}^0 \bm{U}^{0 \top} \mathbf{A}^* \mathbf{B}^*)\\
        =& \mathcal{L}(\mathcal{D}_{\text{train}}; \mathbf{W}^0 + \mathbf{A}^* \mathbf{B}^*),
    \end{align*}
    \end{small}
    where the last equality follows from the invariance assumption. 
    This gives the first inequality:
    \begin{small}
    \begin{align*}
        &\min_{\bm{\Theta}_1 \in \R^{d_2 \times 2r}, \bm{\Theta}_2 \in \R^{2r \times d_2}} \mathcal{L}(\mathcal{D}_{\text{train}}; \mathbf{W}^0 + f(\mathbf{W}^0; (\bm{\Theta}_1, \bm{\Theta}_2))) \\&\leq \mathcal{L}(\mathcal{D}_{\text{train}}; \mathbf{W}^0 + f(\mathbf{W}^0; (\bm{\Theta}_1^\#, \bm{\Theta}_2^\#)))\\
        &= \mathcal{L}(\mathcal{D}_{\text{train}}; \mathbf{W}^0 + \mathbf{A}^* \mathbf{B}^*)\\
        &= \min_{\mathbf{A} \in \R^{d_1 \times r}, \mathbf{B} \in \R^{r \times d_2}} \mathcal{L}(\mathcal{D}_{\text{train}}; \mathbf{W}^0 + \mathbf{A} \mathbf{B}).
    \end{align*}\end{small}

    We next show the following inequality:
    \begin{small}
    \begin{align*}
        &\min_{\mathbf{A} \in \R^{d_1 \times r}, \mathbf{B} \in \R^{r \times d_2}} \mathcal{L}(\mathcal{D}_{\text{train}}; \mathbf{W}^0 + \mathbf{A} \mathbf{B})
        \\&\leq \min_{\bm{\Theta}_1 \in \R^{d_2 \times r}, \bm{\Theta}_2 \in \R^{r \times d_2}} \mathcal{L}(\mathcal{D}_{\text{train}}; \mathbf{W}^0 + f(\mathbf{W}^0; (\bm{\Theta}_1, \bm{\Theta}_2))).
    \end{align*}
    \end{small}
    Take $\mathbf{A}^\# = \sigma(\mathbf{W}^0 \bm{\Theta}_1^*) \in \R^{d_1 \times r}$ and $\mathbf{B}^\# = \bm{\Theta}_2^* \in \R^{r \times d_2}$, where $(\bm{\Theta}_1^*, \bm{\Theta}_2^*) = \argmin_{\bm{\Theta}_1 \in \R^{d_2 \times r}, \bm{\Theta}_2 \in \R^{r \times d_1}} \mathcal{L}(\mathcal{D}_{\text{train}}; \mathbf{W}^0 + f(\mathbf{W}^0; (\bm{\Theta}_1, \bm{\Theta}_2)))$. 
    The conclusion follows from
    \begin{small}
    \begin{align*}
        &\min_{\mathbf{A} \in \R^{d_1 \times r}, \mathbf{B} \in \R^{r \times d_2}} \mathcal{L}(\mathcal{D}_{\text{train}}; \mathbf{W}^0 + \mathbf{A} \mathbf{B})
        \\&\leq \mathcal{L}(\mathcal{D}_{\text{train}}; \mathbf{W}^0 + \mathbf{A}^\# \mathbf{B}^\#)\\
        &= \mathcal{L}(\mathcal{D}_{\text{train}}; \mathbf{W}^0 + \sigma(\mathbf{W}^0 \bm{\Theta}_1^*) \bm{\Theta}_2^*)\\
        &= \min_{\bm{\Theta}_1 \in \R^{d_2 \times r}, \bm{\Theta}_2 \in \R^{r \times d_1}} \mathcal{L}(\mathcal{D}_{\text{train}}; \mathbf{W}^0 + f(\mathbf{W}^0; (\bm{\Theta}_1, \bm{\Theta}_2))).
    \end{align*}
    \end{small}
\end{proof}

\subsection{Theoretical Analysis of {\name} under sinusoid activation function}\label{sec: another theory}

Here we consider a sinusoid activation function $\sigma_\textnormal{p}(x) = \sin(2\pi x)$ \citep{gashler2014training} and design $f(\mathbf{W}^0; \bm{\theta}) = \sigma_\textnormal{p}(\mathbf{W}^0 \bm{\Theta}_1) \bm{\Theta}_2$ with $\bm{\theta} = (\bm{\Theta}_1, \bm{\Theta}_2)$. With this periodic activation function, we can show a stronger result that {\name} has expressivity (almost) greater than or equal to a LoRA with more parameters when $d_1 \gg d_2$.

\begin{proposition}[Expressivity of {\name} with Sine Activation]\label{prop: equivalence sine}
    Suppose that there exists a row of $\mathbf{W}^{0}$, whose entries are linearly independent over the rationals. 
    Then, for any $r > 0$, $\mathbf{A} \in \R^{d_1 \times r}$ and $\mathbf{B} \in \R^{r \times d_2}$, and 
    $\epsilon > 0$, there exists some $\bm{\Theta}_1^* \in \R^{d_2 \times r}$ and $\bm{\Theta}_2^* \in \R^{r \times d_2}$ such that
    \begin{align*}
        \|\mathbf{A} \mathbf{B} - \sigma_\textnormal{p}(\mathbf{W}^{0} \bm{\Theta}_1^*) \bm{\Theta}_2^*\|_{\textnormal{F}} \leq \epsilon.
    \end{align*}
\end{proposition}
Proposition \ref{prop: equivalence sine} shows that the class of updates $\Delta \mathbf{W} = \sigma_\textnormal{p}(\mathbf{W}^{0} \bm{\Theta}_1) \bm{\Theta}_2$ by {\name} with $2r d_2$ parameters is dense in the class of updates $\Delta \mathbf{W} = \mathbf{A} \mathbf{B}$ by LoRA with $r(d_1 + d_2)$ parameters. When $d_2 \ll d_1$, this shows better parameter efficiency of {\name}.

Examining the proof of Proposition \ref{prop: equivalence sine}, it is straightforward to show that the result holds for any continuous and periodic activation function whose range contains an open interval centered at 0.

\begin{proof}
    This proof relies on Kronecker's theorem (Theorem 7.9 in \citet{apostolmodular}) from number theory, which shows that for all $j \in \R^q$, the fractional parts of $(c t_1, c t_2, \dots, c t_q)^\top$ is dense in $[0, 1]^q$ over $c \in \R$, as long as $t_1, \dots, t_q$ are linearly independent over the rationals.

    Let $\mathbf{W}_{j^*}$ be the $j^*$-th column of $\mathbf{W}^{0}$ whose entries are linearly independent over the rationals.
    Since $\mathbf{A} \mathbf{B}$ has a scale ambiguity, we can assume that $\mathbf{A}$ is a matrix whose entries are bounded by 1 without loss of generality. Write $\mathbf{A} = (\mathbf{A}_1, \mathbf{A}_2, \dots, \mathbf{A}_r)$.

    Take $\epsilon' > 0$ whose value will be determined later.
    From Kronecker's theorem, for each $\mathbf{A}_j$ there exists some $c_j \in \R$ such that
    \begin{align*}
        \abs{\{c_j \mathbf{W}_{j^*}\} - \frac{\arcsin(\mathbf{A}_j)}{2\pi}} \leq \epsilon',
    \end{align*}
    where $\{\mathbf{B}\}$ is a vector whose entries are the fractional part of the corresponding entry of $\mathbf{B}$, and $\arcsin$ is applied elementwisely.
    
    Let $\bm{\Theta}_1^* = (c_1 \bm{e}_{j^*}, c_2 \bm{e}_{j^*}, \dots, c_r \bm{e}_{j^*})$, where $\bm{e}_{j^*}$ is the $j^*$-th standard basis vector in $\R^{d_2}$.
    Using the fact that $2\pi\{c_j \mathbf{W}_{j^*}\} = 2 \pi c_j \mathbf{W}_{j^*} \mod 2\pi$, we have
    \begin{align}
        \notag&\norm{\sigma_\textnormal{p}(\mathbf{W}^{0} \bm{\Theta}_1^*) - \mathbf{A}}_{\textnormal{F}}^2 \\\notag&= \norm{\sigma_\textnormal{p}((c_1 \mathbf{W}_{j^*}, c_2\mathbf{W}_{j^*}, \dots c_r \mathbf{W}_{j^*})) - \mathbf{A}}_{\textnormal{F}}^2\\
        &\leq \sum_j \norm{\sin(2 \pi c_j \mathbf{W}_{j^*}) - \mathbf{A}_j}^2 \leq 4 \pi^2 r \epsilon'^2,\label{eq: sine bound}
    \end{align}
    where the last inequality follows from \eqref{eq: sine bound} and the fact that $\sin(x)$ is Lipschitz continuous with Lipschitz constant 1.
    Hence by choosing $\bm{\Theta}_2^* \gets \mathbf{B}$, we have
    \begin{align*}
        &\norm{\mathbf{A} \mathbf{B} - \sigma_\textnormal{p}(\mathbf{W}^{0} \bm{\Theta}_1^*) \bm{\Theta}_2^*}_{\textnormal{F}}^2 \\\leq& \norm{\mathbf{B}}^2 \norm{\sigma_\textnormal{p}(\mathbf{W}^{0} \bm{\Theta}_1^*) - \mathbf{A}}_{\textnormal{F}}^2 \\\leq& 4 \pi^2 \|\mathbf{B}\|^2 r \epsilon'^2.
    \end{align*}
    Choose $\epsilon' = \epsilon / (2 \pi \sqrt{r} \|\mathbf{B}\|)$, then the proof is complete.
\end{proof}

\section{Hyperparameters}
\label{hyper}

We provide the specific hyperparameters used in our experiments to ensure reproducibility. For most of our experiments, we use the standard implementation of {\name}, which we refer to as vanilla {\name}. The neural network architecture in vanilla {\name} consists of only two layers: an input layer and an output layer. We select this approach because vanilla {\name} offers the benefits of simplicity in implementation, a low parameter count, and sufficient adaptation power. Nonetheless, we dedicate Section~\ref{depth-analysis} to exploring more complex adaptation networks and their effect on performance.

\subsection{Image Classification}
\label{cv-hyper}

Hyperparameters for {\name} for Fig. \ref{fig:module-part-analysis} are provided in Table \ref{tab:cv-hyper}. We tune the classification head and the backbone separately and provide detailed settings for each dataset. All weight decay values are not tuned and follow the settings from \citet{fourierft}. The scaling factor $s$ is set to $1.0$. The hidden layer dimension $r$ for MHSA is set to 7 in the QV-setting, while both hidden layer dimensions for MHSA and MLP are set to 2 in the QV-MLP-setting described in Section \ref{cv-settings}.

\begin{table*}[ht]
\centering
\caption{\centering Hyperparameter of image classification for {\name}.}
\label{tab:cv-hyper}
\resizebox{0.8\textwidth}{!}{%
\begin{tabular}{@{}l|clllllll@{}}
\toprule
Hyperparameter & OxfordPets & \multicolumn{1}{c}{StanfordCars} & \multicolumn{1}{c}{CIFAR10} & \multicolumn{1}{c}{DTD} & \multicolumn{1}{c}{EuroSAT} & \multicolumn{1}{c}{FGVC} & \multicolumn{1}{c}{RESISC45} & \multicolumn{1}{c}{CIFAR100} \\ \midrule
Epochs & \multicolumn{8}{c}{10} \\
Optimizer & \multicolumn{8}{c}{AdamW} \\
LR Schedule & \multicolumn{8}{c}{Linear} \\
Weight Decay & \multicolumn{1}{c}{8E-4} & \multicolumn{1}{c}{4E-5} & \multicolumn{1}{c}{9E-5} & \multicolumn{1}{c}{7E-5} & \multicolumn{1}{c}{3E-4} & \multicolumn{1}{c}{7E-5} & \multicolumn{1}{c}{3E-4} &  \multicolumn{1}{c}{1E-4} \\
\specialrule{0em}{1pt}{1pt}
\hline
\specialrule{0em}{1pt}{1pt}
\small QV \\
\specialrule{0em}{1pt}{1pt}
\hline
\specialrule{0em}{1pt}{1pt}
Learning Rate ({\name}) & \multicolumn{1}{c}{5E-3} & \multicolumn{1}{c}{1E-2} & \multicolumn{1}{c}{5E-3} & \multicolumn{1}{c}{1E-2} & \multicolumn{1}{c}{5E-3} & \multicolumn{1}{c}{1E-2} & \multicolumn{1}{c}{5E-3} & \multicolumn{1}{c}{5E-3} \\
Learning Rate (Head) & \multicolumn{1}{c}{5E-3} & \multicolumn{1}{c}{1E-2} & \multicolumn{1}{c}{5E-3} & \multicolumn{1}{c}{1E-2} & \multicolumn{1}{c}{5E-3} & \multicolumn{1}{c}{1E-2} & \multicolumn{1}{c}{1E-2} & \multicolumn{1}{c}{5E-3} \\ 
\specialrule{0em}{1pt}{1pt}
\hline
\specialrule{0em}{1pt}{1pt}
\small QV-MLP \\
\specialrule{0em}{1pt}{1pt}
\hline
\specialrule{0em}{1pt}{1pt}
Learning Rate ({\name}) & \multicolumn{1}{c}{5E-3} & \multicolumn{1}{c}{5E-3} & \multicolumn{1}{c}{5E-3} & \multicolumn{1}{c}{1E-2} & \multicolumn{1}{c}{5E-3} & \multicolumn{1}{c}{5E-3} & \multicolumn{1}{c}{1E-2} & \multicolumn{1}{c}{5E-3} \\
Learning Rate (Head) & \multicolumn{1}{c}{5E-3} & \multicolumn{1}{c}{1E-2} & \multicolumn{1}{c}{5E-3} & \multicolumn{1}{c}{1E-2} & \multicolumn{1}{c}{5E-3} & \multicolumn{1}{c}{1E-2} & \multicolumn{1}{c}{1E-2} & \multicolumn{1}{c}{5E-3} \\ 
\bottomrule
\end{tabular}%
}
\end{table*}

\begin{table}[h]
\centering
\caption{\centering Hyperparameter of commonsense reasoning for {\name}.}
\label{tab:commonsense-hyper}
\begin{tabular}{@{}clcccccccc@{}}
\toprule
Hyperparameter & \multicolumn{8}{c}{Commonsense Reasoning} \\ \midrule
\multicolumn{1}{c|}{Hidden Layer Dimension} & \multicolumn{8}{c}{32} \\
\multicolumn{1}{c|}{$\alpha$} & \multicolumn{8}{c}{32} \\
\multicolumn{1}{c|}{Dropout} & \multicolumn{8}{c}{0.05} \\
\multicolumn{1}{c|}{Optimizer} & \multicolumn{8}{c}{Adam W} \\
\multicolumn{1}{c|}{Learning Rate} & \multicolumn{8}{c}{3e-4} \\
\multicolumn{1}{c|}{Batch Size} & \multicolumn{8}{c}{16} \\
\multicolumn{1}{c|}{Warmup Steps} & \multicolumn{8}{c}{100} \\
\multicolumn{1}{c|}{Epochs} & \multicolumn{8}{c}{1} \\
\bottomrule
\end{tabular}%
\end{table}

\begin{table}[h]
\centering
\caption{\centering Hyperparameter of arithmetic reasoning for {\name}.}
\label{tab:math-hyper}
\resizebox{0.4\textwidth}{!}{%
\begin{tabular}{@{}clcccccccc@{}}
\toprule
Hyperparameter & \multicolumn{8}{c}{Arithmetic Reasoning} \\ \midrule
\multicolumn{1}{c|}{Hidden Layer Dimension} & \multicolumn{8}{c}{64} \\
\multicolumn{1}{c|}{$\alpha$} & \multicolumn{8}{c}{64} \\
\multicolumn{1}{c|}{Dropout} & \multicolumn{8}{c}{0.05} \\
\multicolumn{1}{c|}{Optimizer} & \multicolumn{8}{c}{Adam W} \\
\multicolumn{1}{c|}{Learning Rate} & \multicolumn{8}{c}{3e-4} \\
\multicolumn{1}{c|}{Batch Size} & \multicolumn{8}{c}{16} \\
\multicolumn{1}{c|}{Warmup Steps} & \multicolumn{8}{c}{100} \\
\multicolumn{1}{c|}{Epochs} & \multicolumn{8}{c}{3} \\
\bottomrule
\end{tabular}%
}
\end{table}

\subsection{Natural Language Understanding}
\label{nlu-hyper}

We provide used hyper-parameters for {\name} in natural language understanding on the GLUE benchmark in Table \ref{tab:nlu-hyper} and Table \ref{tab:nlu-hyper-s}. The reported results are obtained when using a depth of 6 for {\name}. The learning rates for the head and the backbone are tuned separately. The scaling factor $s$ is searched in $\{0.01, 0.1, 1.0\}$. For reproducibility, we fix the seed as 0. The hidden layer dimension $r$ is set to 8 in {\name}-L and 1 in {\name}-S. More specifically, we apply {\name} to all layers in RoBERTa-base for {\name}-L, while only applying {\name} to layers $\{4,5,6,7,8,9,10,11\}$ for {\name}-S to reduce the number of trainable parameters. The seed is fixed for reproducibility.

\begin{table*}[t]
\centering
\caption{\centering Hyperparameter of GLUE benchmark for {\name}-L.}
\label{tab:nlu-hyper}
\begin{tabular}{@{}lccccccccc@{}}
\toprule
Hyperparameter & \multicolumn{1}{|c}{STS-B} & \multicolumn{1}{c}{RTE} & \multicolumn{1}{c}{MRPC} & \multicolumn{1}{c}{CoLA} & \multicolumn{1}{c}{SST-2} & \multicolumn{1}{c}{QNLI} & \multicolumn{1}{c}{MNLI} & \multicolumn{1}{c}{QQP} \\ \midrule
\multicolumn{1}{l|}{Optimizer} & \multicolumn{8}{c}{AdamW} \\
\multicolumn{1}{l|}{LR Schedule} & \multicolumn{8}{c}{Linear} \\
\multicolumn{1}{l|}{Learning Rate ({\name})} & 5E-3 & 5E-3 & 5E-3 & 1E-3 & 5E-3 & 1E-3 & 5E-3 & 5E-3 \\
\multicolumn{1}{l|}{Learning Rate (Head)} & 5E-3 & 5E-3 & 5E-3 & 1E-3 & 5E-3 & 1E-3 & 5E-3 & 5E-3 \\
\multicolumn{1}{l|}{Scaling} & 0.1 & 0.01 & 0.01 & 0.1 & 0.01 & 0.01 & 0.01 & 0.01 \\
\multicolumn{1}{l|}{Max Seq. Len} & 512 & 512 & 512 & 512 & 512 & 512 & 512 & 512 \\
\multicolumn{1}{l|}{Batch Size} & 64 & 32 & 64 & 64 & 32 & 32 & 32 & 64 \\  \bottomrule
\end{tabular}%
\end{table*}
\begin{table*}[t]
\centering
\caption{\centering Hyperparameter of GLUE benchmark for {\name}-S.}
\label{tab:nlu-hyper-s}
\begin{tabular}{@{}lccccccccc@{}}
\toprule
Hyperparameter & \multicolumn{1}{|c}{STS-B} & \multicolumn{1}{c}{RTE} & \multicolumn{1}{c}{MRPC} & \multicolumn{1}{c}{CoLA} & \multicolumn{1}{c}{SST-2} & \multicolumn{1}{c}{QNLI} & \multicolumn{1}{c}{MNLI} & \multicolumn{1}{c}{QQP} \\ \midrule
\multicolumn{1}{l|}{Optimizer} & \multicolumn{8}{c}{AdamW} \\
\multicolumn{1}{l|}{LR Schedule} & \multicolumn{8}{c}{Linear} \\
\multicolumn{1}{l|}{Learning Rate ({\name})} & 5E-3 & 1E-3 & 5E-3 & 5E-3 & 5E-3 & 1E-3 & 5E-3 & 1E-3 \\
\multicolumn{1}{l|}{Learning Rate (Head)} & 1E-3 & 1E-3 & 5E-3 & 1E-3 & 5E-3 & 1E-3 & 5E-3 & 1E-3 \\
\multicolumn{1}{l|}{Scaling} & 0.1 & 1.0 & 0.01 & 0.1 & 0.01 & 0.1 & 0.01 & 1.0 \\
\multicolumn{1}{l|}{Max Seq. Len} & 512 & 512 & 512 & 512 & 512 & 512 & 512 & 512 \\
\multicolumn{1}{l|}{Batch Size} & 64 & 32 & 64 & 64 & 32 & 32 & 32 & 64 \\  \bottomrule
\end{tabular}%
\end{table*}

\subsection{Commonsense Reasoning}

We provide hyperparameters settings of {\name} for commonsense reasoning task in Table  \ref{tab:commonsense-hyper}. We follow the hyperparameters settings in MiLoRA \citep{milora}. We limit all samples to a maximum of 256 tokens. For evaluation, we set a maximum token number of 32.

\subsection{Arithmetic Reasoning}

We provide hyperparameters settings of {\name} for arithmetic reasoning task in Table  \ref{tab:math-hyper}. We follow the hyper-parameters settings in MiLoRA \citep{milora}. We limit all samples to a maximum of 2048 tokens. For evaluation, we set a maximum token number of 256 on GSM8K \citep{gsm8k} dataset. On MATH \citep{MATH}, we set the maximum new token to 512.
\section{Datasets}
\label{datasets}

In this section, we provide a detailed description of the datasets used in our experiments.

\subsection{Image Classification}
\label{cv-dataset}

For image classification, we provide detailed information about the used datasets  in Table \ref{tab:cv-hyper}.

\begin{table}[h!]
\centering
\caption{\centering Detailed information of image classification tasks.}
\label{tab:cv-data}
\begin{tabular}{@{}l|lrrrc@{}}
\toprule
Dataset & \#Class & \#Train & \#Val & \#Test & Rescaled resolution \\ \midrule
OxfordPets & 37 & 3,312 & 368 & 3,669 & \multirow{8}{*}{$224\times224$} \\
StandfordCars & 196 & 7,329 & 815 & 8,041 &  \\
CIFAR10 & 10 & 45,000 & 5,000 & 10,000 &  \\
DTD & 47 & 4,060 & 452 & 1,128 &  \\
EuroSAT & 10 & 16,200 & 5,400 & 5,400 &  \\
FGVC & 100 & 3,000 & 334 & 3,333 &  \\
RESISC45 & 45 & 18,900 & 6,300 & 6,300 &  \\
CIFAR100 & 100 & 45,000 & 5,000 & 10,000 &  \\ \bottomrule
\end{tabular}%
\end{table}

\subsection{Natural Language Understanding}
\label{nlu-dataset}

The GLUE benchmark comprises 8 NLP datasets: MNLI, SST-2, MRPC, CoLA, QNLI, QQP, RTE, and STS-B, covering tasks such as inference, sentiment analysis, paraphrase detection, linguistic acceptability, question-answering, and textual similarity. We provide detailed information about them in Table \ref{tab:nlu-data}.

\begin{table}[h!]
\centering
\caption{\centering Detailed information of the GLUE benchmark. STS-B is a regression task, while all other tasks are either single-sentence or sentence-pair classification tasks.}
\label{tab:nlu-data}
\begin{tabular}{@{}cccccccc@{}}
\toprule
\multicolumn{1}{l|}{\textbf{Corpus}} & \textbf{Task} & \textbf{Metrics} & \textbf{\# Train} & \textbf{\# Val} & \textbf{\# Test} & \textbf{\# Labels} \\ \midrule
\multicolumn{7}{c}{Single-Sentence Tasks} \\ \midrule
\multicolumn{1}{l|}{CoLA} & Acceptability & Matthews Corr. & 8.55k & 1.04k & 1.06k & 2 \\
\multicolumn{1}{l|}{SST-2} & Sentiment & Accuracy & 67.3k & 872 & 1.82k & 2 \\ \midrule
\multicolumn{7}{c}{Similarity and Paraphrase Tasks} \\ \midrule
\multicolumn{1}{l|}{MRPC} & Paraphrase & Accuracy/F1 & 3.67k & 408 & 1.73k & 2 \\
\multicolumn{1}{l|}{STS-B} & Sentence similarity & Pearson/Spearman Corr. & 5.75k & 1.5k & 1.38k & 1 \\
\multicolumn{1}{l|}{QQP} & Paraphrase & Accuracy/F1 & 364k & 40.4k & 391k & 2 \\ \midrule
\multicolumn{7}{c}{Inference Tasks} \\ \midrule
\multicolumn{1}{l|}{MNLI} & NLI & Accuracy & 393k & 19.65k & 19.65k & 3 \\
\multicolumn{1}{l|}{QNLI} & QA/NLI & Accuracy & 105k & 5.46k & 5.46k & 2 \\
\multicolumn{1}{l|}{RTE} & NLI & Accuracy & 2.49k & 277 & 3k & 2 \\ \bottomrule
\end{tabular}%
\end{table}

\subsection{Commonsense Reasoning}
\begin{table}[h!]
\centering
\caption{\centering Detailed information of commonsense reasoning task.}
\label{tab:commonsense_data}
\begin{tabular}{@{}l|lccc}
\toprule
Dataset & \#Class & \#Train & \#Dev& \#Test \\ \midrule
BoolQ& Binary classification& 9,427& 3,270& 3,245\\
PIQA& Binary classification& 16,113& 1,838& 3,000\\
SIQA& Ternary classification& 33,410& 1,954& 2,224\\
HellaSwag& Quaternary classification& 39,905& 10,042& 10,003\\
WinoGrande& Binary classification& 40,398& 1,267& 1,767\\
ARC-e& Quaternary classification& 2,251& 570& 2,376\\
ARC-c& Quaternary classification& 1,119& 229& 1,172\\
OBQA& Quaternary classification& 4,957& 500& 500\\ \bottomrule
\end{tabular}%
\end{table}
\begin{table}[h!]
\centering
\caption{\centering Detailed information of arithmetic reasoning task.}
\label{tab:math_data}
\begin{tabular}{@{}l|ccc}
\toprule
Dataset & \#Train & \#Dev& \#Test \\ \midrule
GSM8K& 7,473& 1,319& 1,319\\
MATH& 12,500& 500& 5,000\\ \bottomrule
\end{tabular}%
\end{table}
For commonsense reasoning task, we use 8 datasets, including BoolQ, PIQA, SIQA, HellaSwag, WinoGrande, ARC-e, ARC-c and OBQA. The detailed information is provided in Table \ref{tab:commonsense_data}.

\subsection{Arithmetic Reasoning}

Detailed information for arithmetic reasoning task is provided in Table \ref{tab:math_data}. GSM8K consists of high quality grade school math problems, typically free-form answers. MATH includes classifications from multiple mathematical domains, such as algebra, counting\_and\_probability, geometry, intermediate\_algebra, number\_theory, prealgebra and precalculus.

\appendix
\end{document}